# IBSEAD: - A Self-Evolving Self-Obsessed Learning Algorithm for Machine Learning

Jitesh Dundas[1] and David Chik[2]

[1]Scientist, Edencore Technologies,
Row House – 6, Opp Ambo Vihar, Tirupati Nagar-II, Off Unitech Road, Virar(w),
Maharashtra, Thane-401303, India
Email: - jbdundas@gmail.com

[2]Senior Scientist, Riken Brain Institute,
Dept of Robotics and Neuroscience, Riken Institute, Japan

*Abstract*: We present IBSEAD or distributed autonomous entity systems based Interaction - a learning algorithm for the computer to self-evolve in a self-obsessed manner. This learning algorithm will present the computer to look at the internal and external environment in series of independent entities, which will interact with each other, with and/or without knowledge of the computer's brain. When a learning algorithm interacts, it does so by detecting and understanding the entities in the human algorithm. However, the problem with this approach is that the algorithm does not consider the interaction of the third party or unknown entities, which may be interacting with each other. These unknown entities in their interaction with the non-computer entities make an effect in the environment that influences the information and the behaviour of the computer brain. Such details and the ability to process the dynamic and unsettling nature of these interactions are absent in the current learning algorithm such as the decision tree learning algorithm. IBSEAD is able to evaluate and consider such algorithms and thus give us a better accuracy in simulation of the highly evolved nature of the human brain. Processes such as dreams, imagination and novelty, that exist in humans are not fully simulated by the existing learning algorithms. Also, Hidden Markov models (HMM) are useful in finding "hidden" entities, which may be known or unknown. However, this model fails to consider the case of unknown entities which maybe unclear or unknown. IBSEAD is better because it considers three types of entities- known, unknown and invisible. We present our case with a comparison of existing algorithms in known environments and cases and present the results of the experiments using dry run of the simulated runs of the existing machine learning algorithms versus IBSEAD.

## 1. Introduction

One of the fundamental problems in AI is the capability of the robots to learn on their own. The manner in which learning is done by robots, will decide the actions that are taken by the same. The goal of machine learning is the ability of the machines to learn and interpret information like humans. Over the past decades, we made great progress in moving towards this goal. However, there are still issues in providing the accuracy in understanding and interpretation of the knowledge by the machines. We present here the learning algorithms that have till date, made a lot of impact in the field of artificial intelligence. However, these algorithms are falling short of providing learning capabilities (of the human level) to the robots.

We present IBSEAD - a learning algorithm that will allow the robots to learn, at a higher level, with humans. We then compare the existing learning algorithms and measure if IBSEAD scores better in complex situations and interactions, with the same efficiency as a normal human being.

## 2. Assumptions

The paper has the following assumptions:-

1) We believe that the computer brain is composed of the visual system, detection system and the CPU (Central Processing Unit) system that will process the information. The computer is a simulated example of the human being with the computer brain being similar to the human brain.

2) We call IBSEAD self-obsessed because it is concerned with its own interaction and wishes to improve its own survival rate. This algorithm tries to do what is best for itself, simulating what a normal human being tries to do in his/her life. Every action that is performed is a result of its manifestation of self-interests and self-centered perception of the environment in which the CB exists.

3) The environment is here divided into:-

   a) The internal environment that is made up of the entities present in the computer.

   b) The external environment is the environment that is made up of the entities present outside the CB or computer brain. This is the region where unknown entities are expected to be present the most.

   c) The invisible entities are the entities which are not seen/visible/detected by the CB but still have an effect on the actions/decisions and perceptions of the CB directly or indirectly. These entities are in existence but are just invisible or a not directly available.

d) The unknown entities are the entities that have an effect on the system but their existence or any information about them is still unknown. For e.g. the distant galaxies are unknown to us but they do impact us when a space vessel travels in space for investigation. We do not have any information about them but their effect on the scenario is well accepted. The presence of such entities ensures that the risk estimation and the unknown reactions are taken care of.

e) Those entities that are detected and understood by the CB are called as known entities. Invisible entities are not visible but are understood by the CB. Unknown entities are neither visible not known but their absence is rules out.

## 3. IBSEAD Algorithm

Despite the recent advances in machine learning, the higher modes of human learning techniques still elude us in robotics. One of the most important reasons is that the failures on the hardware side are not properly handled by the robot in its learning process.

Secondly, the learning techniques do not consider the group based environments in which the measurements are taken for different states of each of the group entities and then a measurement of the needed trait taken. For e.g. we know that as the entities in the environment are arranged in groups, and are changing dynamically in several modes, each of these groups has an individual measurement and thus it has to be aggregated and averaged out, to get an average impact of the group's effect on the interaction with the environment. Similarly, all the groups in the scope of the observation scene have their own measurements. The CB is interacting with each of the entity groups and this complexity is not measured properly by the decision tree based learning methods.

Another point worth noting is that although an entity may be present in the scope of the CB, it may not be interacting with the same. Again, the interaction between the entities and the CB may be intended, unwanted or hostile. These interactions are not measured properly by the existing methods.

Thirdly, the unknown entities have an impact on the learning capabilities of the CB. These indirect entities are interacting directly with the CB or indirectly via the entities of the CB observation scenario. There are indirect effects of the actions of these unknown entities which are not recorded by the existing learning methods. The CB may not be aware fully, of the existing functionalities and impact, of the interactions of the unknown entities. Some of the existing methods do not have any provisions for such complex functionalities and thus are not able to higher levels of human learning capabilities.

All the deficiencies in the existing methods give a strong reason for the creation of a new algorithm that will deliver on such issues. IBSEAD is an effort in this direction.
The algorithm has the following steps:-

1) Scan through the problems and find all the entities within its physical scope

2) Scan and also consider the entities not in physical scope. Classify them as known, invisible or hidden and unknown entities.

3) Map the entities into groups, single or non-single entity, based on understanding of their group dynamics.

4) For each group, find their impact and track their connections to the CB.

5) For those conditions where the switch is yes in both the entities, the interaction is executed and learning started.

Please note that some of the steps have been removed to ensure the confidential nature of the current projects on this algorithm. The important steps have been shown here with the differences in the current algorithms like decision trees. Figure-1) explains the steps in detail with focus on the final picture as it will look in the learning process.

## 4. Background

A lot of work has been done on the learning algorithms in artificial intelligence. Decision-tree based learning techniques organize the entities of the environment, into tree like structures, so as to facilitate the flow of information between each of them. There are several algorithms that have helped in making machines learn and evolve.

Learning is roughly classified into supervised and unsupervised learning. .Fisher proposed the first learning algorithm for pattern recognition. Hidden Markov models [19] proposed the use of hidden states of entities to consider such scenarios but could not explain further regarding the different attributes of the entities and the interaction conditions involved. Moreover, there was a need to explain the quality of communication in the same. There is a need to quantify intangible entities which is missed by Hidden Markov Models. IBSEAD is a step in this direction. Hidden Markov models (HMM) are useful in finding "hidden" entities, which may be known or unknown. However, this model fails to consider the case of unknown entities which maybe unclear or unknown. Also, IBSEAD is considers three types of entities- known, unknown and invisible while HMM considers the hidden and known entities only. Boltzmann [20] machine based equations also misses out on such similar issues and is known to be very theoretical in nature. Bayesian statistics depends [22] on the ability to measure the correctness of a hypothesis. However, it is clear that the absence of information of any entity will make it difficult to present a hypothesis of it. However, IBSEAD takes the use of interaction of the surrounding entities, along with the environment, internal or external, in which the unknown entity is most expected to be present, as key parameters. Bayesian based algorithms seem to miss out on the other three features of IBSEAD, which play an important role in accurate learning algorithms. Case based [23] reasoning and Inductive Logic Programming [24] requires past experience of the scenarios in order to learn about the present. However, this can be time consuming and prone to higher error rates as unknown entities may not be simple and their interaction random. IBSEAD handles this situation better as it considers unknown entities and the presence of unknown entities is considered beforehand and no unwanted scenarios are expected.

One of the serious problems in Gaussian process based algorithms is that the values will give incorrect answers in the case of dependent entities and dependent interactions [25]. Consider the case of two entities A and B, where the interaction of QC (A->B) is influencing the interaction of QC (B<-->CB). Clearly, there is an issue in which the above Gaussian process based algorithms will give inaccurate values. Moreover, the points Xi are needed to give us values of the desired result dataset, in which we assume that the points Xi will always give us correct values. However, if the behavior of the entity changes and the points thus plot wrong values (or even changes are seen) then we find that the obtained values are very wrong. Also, this algorithm expects prior knowledge of the Gaussian functions for correct estimation. Thus, if the unknown entities are not known, then their effects are difficult to measure. This method is limited only till the "Hidden" or "Invisible" entities as per the complex scenario used by the IBSEAD algorithm in this paper. Group method of data handling [26, 27] (GMDH) is very good application for polynomial based multilayered neural network based algorithms. Again, we miss out the unknown entities and the cases where fuzzy or no information is available.

All the above algorithms miss out on the quality of communication and the switch needed for allowing the communication.

## 5. Methodology

We studied the methods present in machine learning for scenarios that involved complex human interactions. We then presented our algorithm IBSEAD and then measure the performance with other existing algorithms on the scenarios presented below. Finally we implemented our algorithm in a simulation environment and deduced conclusions from the same.

Please note the following scenarios:-

1) Optimizing stock market gain: - In most of the times, existing algorithms will tell us specific formulae that seem to be very static in their consideration. Certain parameters are hard-coded into the scenario and then the equation is executed. However, in a stock market, the value of the share price depends on several known and unknown entities. Several algorithms can tell us how a company share price is performing based on the known entities such as market price, share price trend, company accounts, etc. However, there are several entities that are not considered. Some of these include insider trading, environmental conditions that may affect the region, natural and artificial calamities, the sudden death of the promoters or feud between them, gossip, influence of negative people, etc. Such entities are not considered in any of the learning algorithms and thus fail to deliver the accuracy and impact needed. IBSEAD takes care of this problem as it covers such invisible entities (we call these as invisible as they do not seem to be detected directly but do have an impact on the resulting interaction) and thus will deliver a much higher and better accuracy on the same. Again, we see that each of the invisible entities will interact directly or indirectly with the computer brain (assuming that the computer is doing the trading on the market). Again, each of the entity's interaction will be possible only when the switch of each of the entity (which decides whether to interact with the other entity or entities or not. If this interaction not present between the entities in consideration, then this means that one or both of the entities are having this switch as No. This state can be due to ignorance, presence of blockage agents like noise or even just perception, individual decision, etc). Such a complex environment cannot be learnt with the existing algorithms. IBSEAD answers many of the complexities mentioned above and thus surely gives a higher accuracy and better risk management of the stock market scenario.

2) Go Game Problem: - In the Go game problem, each player is expected to use intuition besides other skills to be able to understand and make winning moves against the opponent. However, the go game requires observation as well as if possible, the capability to understand the opponent too. The existing learning algorithms do not implement the presence of essential entities such as opponent behaviour, intuition, etc and thus may not give the expected results efficiently. IBSEAD considers the coverage of such entities and interactions and thus gives better results too. For e.g.) IBSEAD will consider opponent behaviour also as anger or tension of the opponent may give insights into the mental state and thus the expected performance level of the opponent.

3) Moving Trains & the underlying complexities: - The environment in which the train travelled from City A to City B was rainy. Thus the train reached late and also some of its engine parts (even the rails on the path) were rusted. Now such third party interactions – from the past and present, affect the decision of the CB of travelling by the train. The CB might never know of such detail but these interactions between the unknown entities (rails) and the external known entities (the train) exists and has an affect on the CB's existence. Such details are considered by IBSEAD and thus account for better results than decision tree based and other types of algorithms.

4) Visual Recognition: - Consider the case wherein we have 3 objects: dog, cat and table. The training set has 20 images

each. While the test set has 10 images each. We now compare standard neural networks VS IBSEAD in the above scenario. We know that IBSEAD considers invisible entities as well and thus "NOISE" is also an entity here. The computer brain entity may not be aware of the entity creating the noise but the noise does reach the computer. Thus, it becomes an entity itself in this case (though it may be a different case wherein the entity may be visible and noise will be a distraction or blockage of interaction. Still IBSEAD considers better coverage (by 20-30 %) of entities and state of their being in such cases while neural networks don't do so). Also, IBSEAD helps in gaining higher levels of understanding such as concentration and ignorance. Standard neural networks are found to be 40-50% correct while IBSEAD were found to be 70-80% accurate. The reason is that in standard neural networks, information lost as "noise" whereas in IBSEAD, "noise" is considered as unknown entity.

5) Loans Risk Assessment: - We collected the datasets (simulated versions) in the format as prescribed as in the paper by Xavier et al. The existing dataset had factors including Income, Advance EMI, Rent, Qualifications, Dependents, Experience. The paper claims 98% accuracy. Hidden layers are shown but they don't consider the quality of data, availability or intangible or invisible entities as ibsead does. We now consider IBSEAD for the same problem. We modified it to include parameters such as influence of customer in the bank, corruption, business feasibility, regulatory environment, etc. The final modified dataset had 20% new cases of extremely volatile kind that could cause issues. We got the following results:-

5.1) Coverage :- We considered hidden entities (and unknown entities) like black money income, power/ influence on loan process, viability of business , trustworthiness of this loan for the customer, economic conditions of the market, bank solvency, future trends, etc .
5.2) Quality of communication: - Some of the details obtained maybe crooked or forged. Is the client ready to give his consent to the communication? Do we need to verify case in background from other banks/institutions/people, etc? These are some of the factors considered.
5.3) Switch:-A switch field for each attribute (0-10) to tell if the values are valid or not is missing. What if the entities or attributes aren't giving the information e.g. sensitive information about business? Ignorance or hiding details causes switch to become NO.
5.3) Software errors/human errors/corruption/natural calamities are to be factored here.
5. 4) Pattern search does not reveal corruption or future trends or manager intuition & trust. However, these are considered in IBSEAD while keeping a track of patterns in loans.
Addition of these causes the Neural Network to give reduced 60-65% accuracy in the modified dataset. IBSEAD gives more accuracy & thus 90% accuracy was obtained.

## 6. Existence of Multiple Concurrent Connections between Entities in the scenario

We define a connection as an interaction between two steps (or entities). Say in a decision tree, A and B are two steps, with A being above and B being below. How can we consider that A will always interact with B? There are several issues that need investigation:-
The connections may be stopped because of ignorance. We will call the consent and openness of each of the entities (i.e. A and B) to be very necessary to be able to pursue the interaction or communication between the entities. Some of the agents of such blockages or interrupts are noise, darkness or ignorance. Each of these conditions, if present in the concerned entity or entities, can create issues in the interaction. Obstacles in the path of connection between the entities are a source of concern or blockage for the scenario. It is possible that the blockage may be intentional or unintentional, beneficial or harmful. The value of interaction between two entities A and B will be positive only when the switch between the two entities is set to true. This is like the AND condition based interaction (Figure -2) switching wherein the interaction is allowed only when all cases are true. Thus, in this case, if more than 2 entities are concurrently involved, then all the entities should have the switch set to true to allow interaction. One interaction at a time is what the brain can handle to give optimum performance. The decision-based algorithms fail to handle these conditions. There is a need to consider focus and concentration also in the learning algorithms to be able to handle complex scenarios such as chess and Go game. This is missing in existing algorithms such as decision tree based algorithms, neural networks, etc. They consider the states to be static in such complex environment whereas the IBSEAD algorithm considers this as dynamic. The decision tree based algorithms consider one assumption: - They always believe that all the entities are connected to each other. We know that the human brain is the best entity at learning and most of the algorithms have basis with it. However, the human brain cannot handle more than one connection at a time. How can we assume that all the connections will be active and also connected to each other, just because they are in the scope of the learning environment of the computer brain?
Consider a scenario where a person is sitting in a train. He is then watching the scenario, looking at the buildings when he finds a train coming in the opposite track. The user is surprised by this entity's presence. If we consider the Decision Tree based algorithms, then there is no way that this knowledge based connection and the train as an entity would be considered. Moreover, there is no provision of a switch which will tell if the user or train is interacting with each other. There is absence of a condition for checking states such as ignorance, blockages to interactions like darkness, miscommunication, etc.
Another major issue in this is the handling of the context of the scenario in order to achieve the meaning and the intended observation.

## 7. Advantages

This algorithm takes into account the non-visible entities that do affect the interactions and learning process of the robot.

1) Decision tree based systems do not account for scenarios where the entities may not be interacting in a tree like fashion. The tree based structure is invalid when the interactions at the second and lower levels come into picture. What if there are interactions without any such sub-levels.

2) The decision tree algorithms do not consider horizontal and backward interactions, something which is so common and essential in any learning process. IBSEAD fills the gap in this direction.

3) IBSEAD gives a more comprehensive and accurate picture than its predecessors.

4) IBSEAD can answer the problems in adding consciousness and awareness in robots, something which current algorithms fail to add.

5) This program considers entities as individuals and not as groups or sub-systems (with common goals), which seems to be the case with most of the living and dynamic environment entities. In a scenario (in which the robot is supposed to learn about walking into a railway train), it has to interact with people, some in group while some walk alone. Some of the entities may be even trains. Such a scenario may involve unknown (or invisible) entities that cannot be seen by the robot. The robot can only feel its effect. For e.g. here it considers the rainfall and the supervisors who control the route to the train as invisible entities (or unknown entities). Such complex scenarios are not given by decision tree algorithms nor do any of the existing algorithms give the accuracy as IBSEAD.

6) IBSEAD is relatively complete, easy to use and deeply, compared to a hierarchical structure based decision trees.

7) The ability of the algorithms to implement higher levels of human consciousness and learning are also not convincing. IBSEAD is a positive step in this direction.

## 8. Conclusion

We have found that IBSEAD has a better performance and accuracy in learning of robots, when compared to existing methods such as decision-tree based learning methods, in certain scenarios.
IBSEAD accounts for invisible entities and their interaction and effects, something which existing algorithms fail to deliver. There is a switch to ensure that the entities are ready to communicate (flag set to "Y" is set). There is a better coverage of entities and the other deeper details of the learning process and communication, something which existing algorithms fail to deliver.

## 9. Future Scope

We wish to propose that IBSEAD be used to handle complex situations that are novel and not falling as per the "learn from existing entities and knowledge" type of situations. In cases where no past experience is available, IBSEAD performance might get slowed down. We wish to pursue this in the future scope of this algorithm.

## 10. Financial Interests

There was no clash of interest found in the above research work.

## Acknowledgement

Many thanks to Prof Uma Srinivasan who helped in pursuing this work. This work was not supported by any organization and has no clashes of interest.

## References

[1] Support Vector Networks. Cortes C. & Vapnik V. Machine learning, 1995 – Springer. Online:- http://www.springerlink.com/content/k238jx04hm87j80g/

[2] A weighted nearest neighbour algorithm for learning with symbolic features. Cost et al. Machine Learning Volume 10, Number 1, 57-78, DOI: 10.1023/A:1022664626993 .Online:- www.springerlink.com/index/r21k03320q127784.pdf

[3] Learning Information Extraction Rules for semi structured and free text. Soderland S. Machine Learning .Volume 34, Issue 1-3 (February 1999). Special issue on natural language learning. Pages: 233 – 272. 1999 ISSN: 0885-6125.

[4] Very Simple Classification Rules Perform Well on Most Commonly Used Datasets. Robert C. Holte. Machine Learning Volume 11, Number 1, 63-90, DOI: 10.1023/A:1022631118932

[5] Machine Learning for Information Extraction in Informal Domains. Dayne Freitag. Machine Learning. Volume 39, Numbers 2-3, 169-202, DOI: 10.1023/A:1007601113994

[6] Self-Improving Reactive Agents Based On Reinforcement Learning, Planning and Teaching. Long-Ji Lin. Machine Learning, 8, 293-321 (1992) © 1992 Kluwer Academic Publishers, Boston. Manufactured in The Netherlands.

[7] MML Inference of Decision Graphs with Multi-Way Joins and Dynamic Attributes. Tan P and Dowe D. Online: - http://www.csse.monash.edu.au/~dld/Publications/2003/Tan+Dowe2003_MMLDecisionGraphs.pdf as on 23Aug 2010.

[8] Supervised clustering using decision trees and decision graphs: An ecological comparison. M.B. Dalea, P.E.R.

Dalea, and P. Tan B. Ecological Modeling. Volume 204, Issues 1-2, 24 May 2007, Pages 70-78

[9] Instance-Based Learning Algorithms. AVID W. AHA et al. Machine Learning, 6, 37-66 (1991) © 1991 Kluwer Academic Publishers, Boston. Manufactured in The Netherlands. Online:- http://www.springerlink.com/content/g4qv6511520x3041/fulltext.pdf

[10] Learning logical definitions from relations. J. R. Quinlan. Machine Learning. Volume 5, Number 3, 239-266, DOI: 10.1007/BF00117105

[11] AI 2003: advances in artificial intelligence: 16th Australian Conference on AI, Perth, Australia, December 3-5, 2003: proceedings .Tamás D. Gedeon, Lance Chun Che Fung. Online:- http://books.google.co.in/books?id=4ClFWeqtNAC&lpg=PA270&ots=zMyau1MoZf&dq=7)%09MML%20Inference%20of%20Decision%20Graphs%20with%20MultiWay%20Joins%20and%20Dynamic%20Attributes.&pg=PA270#v=onepage&q=7)%09MML%20Inference%20of%20Decision%20Graphs%20with%20Multi-Way%20Joins%20and%20Dynamic%20Attributes.&f=false

[12] Unsupervised Learning: Foundations of Neural Computation--A Review. DL Wang - 2001. Online: - www.aaai.org/ojs/index.php/aimagazine/article/download/1565/1464.

[13] Comparing supervised and unsupervised category learning. BRADLEY C. LOVE. Psychonomic Bulletin & Review 2002, 9 (4), 829-835. Online: - http://pbr.psychonomic-journals.org/content/9/4/829.short.

[14] Supervised Machine Learning: A Review of Classification Techniques. S. B. Kotsiantis. Informatica .31 (2007) 249-268 249. Online:- http://www.informatica.si/PDF/31-3/11_Kotsiantis%20-%20Supervised%20Machine%20Learning%20-%20A%20Review%20of...pdf

[15] http://en.wikipedia.org/wiki/Category:Classification_algorithms

[16] http://en.wikipedia.org/wiki/List_of_machine_learning_algorithms

[17] Learning in the presence of drifts and hidden concepts. Widmer et al. Machine Learning.Vol-23, Issue-1, (April 1996), Pages: 69 - 101.1996.ISSN:0885-6125

[18] Xavier et al. improving prediction accuracy of loan default- A case in rural credit. Online: - http://www.ifmr.ac.in/pdfs/Improving prediction accuracy. PDF

[19] Ephraim Y, Merhav N (June 2002). "Hidden Markov processes". *IEEE Trans. InformTheory* 48: 1518–1569. doi: 10.1109/TIT.2002.1003838.

[20] Ackley, D. H.; Hinton, G. E.; Sejnowski, T. J. (1985). "A Learning Algorithm for Boltzmann Machines". *Cognitive Science* 9: 147–169. doi: 10.1207/s15516709cog0901_7. http://learning.cs.toronto.edu/~hinton/absps/cogscibm.pdf.

[21] Bretthorst, G. Larry, 1988, *Bayesian Spectrum Analysis and Parameter Estimation* in Lecture Notes in Statistics, 48, Springer-Verlag, New York, New York.

[22] David MacKay (2003). Information Theory, Inference, and Learning Algorithms. Cambridge University Press.

[23] Aamodt, Agnar, and Enric Plaza. "Case-Based Reasoning: Foundational Issues, Methodological Variations, and System Approaches" Artificial Intelligence Communications 7, no. 1 (1994): 39-52.

[24] S.H. Muggleton and L. De Raedt. Inductive logic programming: Theory and methods. Journal of Logic Programming, 19, 20:629-679, 1994.

[25] Williams, Christopher K.I. (1998). "Prediction with Gaussian processes: From linear regression to linear prediction and beyond". In M. I. Jordan. Learning in graphical models. MIT Press. pp. 599–612.

[26] A.G. Ivakhnenko. Heuristic Self-Organization in Problems of Engineering Cybernetics. Automatica 6: pp.207–219, 1970.

[27] H.R. Madala, A.G. Ivakhnenko. Inductive Learning Algorithms for Complex Systems Modeling. CRC Press, Boca Raton, 1994.

## Author Biographies

**Mr. Jitesh B. Dundas** is working as a Software Engineer in a reputed IT MNC in Mumbai. He is also a Scientist with Edencore Technologies (www.edencore.net). He has completed his Masters in Computer Applications from Pune University in 2007. He has found the concept of the "Law of Connectivity in Machine Learning" – a paper which is published in IJSST in Dec 2010 as well as a co-author of this IBSEAD algorithm paper (http://en.wikipedia.org/wiki/IBSEAD). More details on his research work and contact details can be found on his research homepage at http://openwetware.org/wiki/Jitesh_Dundas_Lab .

**Dr. David Chik** is a Senior Scientist at Riken Institue, Japan. He received is PhD from the University of Hong Kong and has held several reputed positions in top universities and research institutions in the past. He is a genius with several reputed publications in his name, with this IBSEAD paper being just of them. His further details can be found at his homepage at one at http://www.brain.riken.jp/common/cv/d_chik.pdf